\documentclass[10pt,twocolumn,letterpaper]{article}

\usepackage[pagenumbers]{cvpr} 

\usepackage{graphicx}
\usepackage{amsmath}
\usepackage{amssymb}
\usepackage{booktabs}
\usepackage{multirow}
\usepackage{color}
\usepackage[table]{xcolor}
\usepackage[pagebackref,breaklinks,colorlinks]{hyperref}

\usepackage{float}

\usepackage[capitalize]{cleveref}
\crefname{section}{Sec.}{Secs.}
\Crefname{section}{Section}{Sections}
\Crefname{table}{Table}{Tables}
\crefname{table}{Tab.}{Tabs.}

\definecolor{citecolor}{RGB}{34,139,34}
\definecolor{grayDark}{gray}{0.95}
\definecolor{grayLight}{gray}{0.98}
\definecolor{darkgreen}{rgb}{0.10, 0.55, 0.10}

\newcommand\green[1]{\textcolor{darkgreen}{#1}}

\newcommand{\TableEntry}[2]{\textbf{#1}~\footnotesize{\green{(+#2)}}}
\newcommand{\TableEntryGray}[2]{\textbf{#1}~\footnotesize{{(+#2)}}}

\def\cvprPaperID{4500} 
\def\confName{CVPR}
\def\confYear{2023}

\newcommand{\et}{\emph{et al.}}
\newcommand{\myparagraph}[1]{{ \noindent \bf #1}}
\newcommand{\myparagraphonept}[1]{{ \vspace{1pt} \noindent \bf #1}}

\DeclareMathOperator*{\argmax}{arg\,max}

\begin{document}

\title{Learning to Detect and Segment for Open Vocabulary Object Detection}

\author{Tao Wang\\
Sichuan University\\
{\tt\small twangnh@gmail.com}
\and
Nan Li\\
University of California San Diego\\
{\tt\small nanlucsd@gmail.com}
}
\maketitle

\begin{abstract}

Open vocabulary object detection has been greatly advanced by the recent development of vision-language pre-trained model, which helps recognize novel objects with only semantic categories. The prior works mainly focus on knowledge transferring to the object proposal classification and employ class-agnostic box and mask prediction. In this work, we propose \emph{CondHead}, a principled dynamic network design to better generalize the box regression and mask segmentation for open vocabulary setting. 
The core idea is to conditionally parameterize the network heads on semantic embedding and thus the model is guided with class-specific knowledge to better detect novel categories.
Specifically, CondHead is composed of two streams of network heads, the dynamically aggregated head and dynamically generated head.
The former is instantiated with a set of static heads that are conditionally aggregated, these heads are optimized as experts and are expected to learn sophisticated prediction.
The latter is instantiated with dynamically generated parameters and encodes general class-specific information.
With such a conditional design, the detection model is bridged by the semantic embedding to offer strongly generalizable class-wise box and mask prediction.
Our method brings significant improvement to the state-of-the-art open vocabulary object detection methods with very minor overhead, \eg, it surpasses a RegionClip model by
3.0 detection AP on novel categories, with only 1.1\% more computation.

\end{abstract}

\section{Introduction}
\label{sec:intro}

\begin{figure}[t]
	\begin{center}
	\includegraphics[width=1\linewidth]{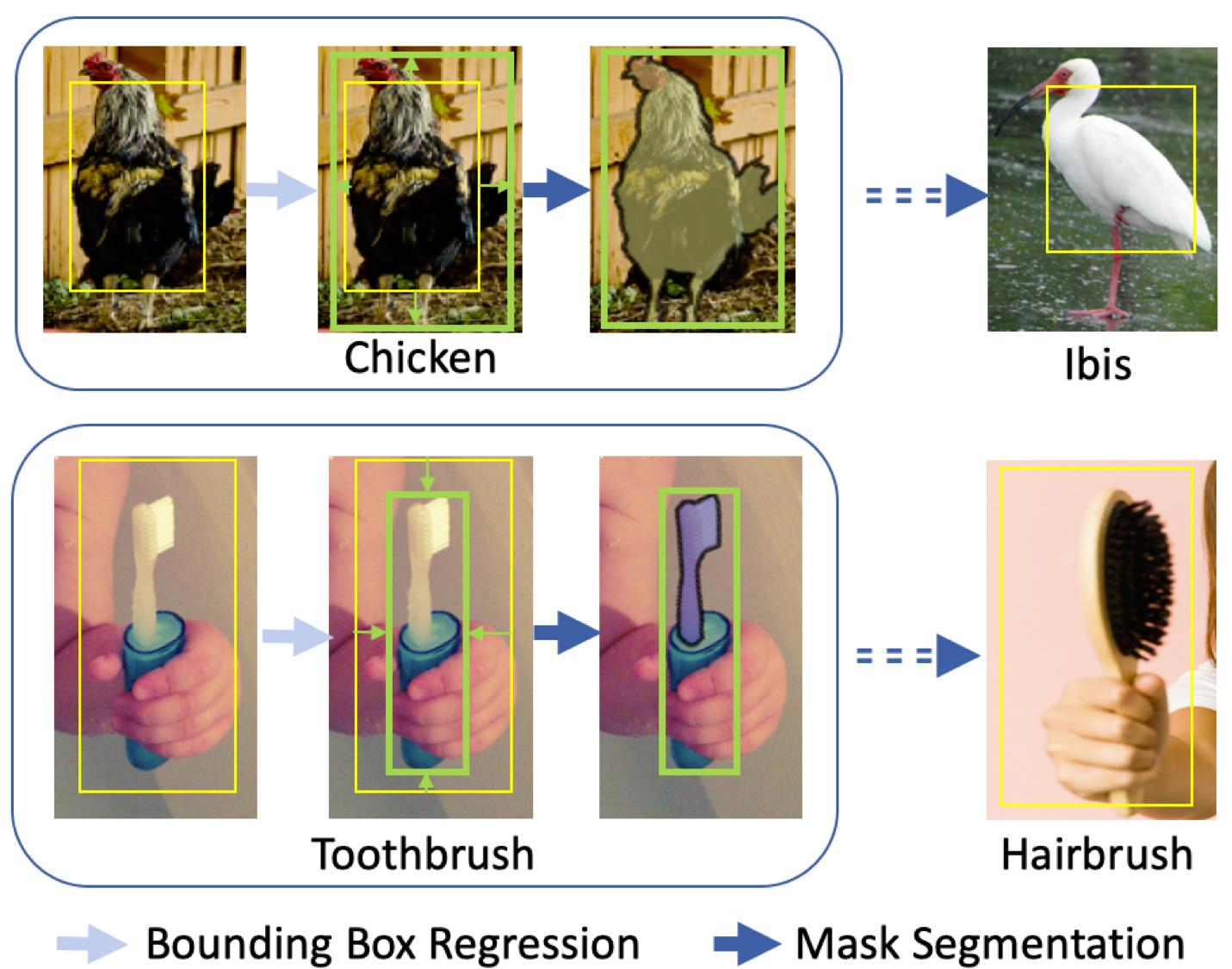}
	\end{center}
	\caption{Illustration of our main intuition. Given the object proposals, the bounding box regression and mask segmentation learned from some object categories could generalize to the target category. For example, the knowledge learned from a chicken could help detect and segment the long thin feet and the small head of an ibis (upper row). Similarly for the hairbrush, the knowledge learned from the toothbrush could better handle the extreme aspect ratio and occlusion from the hand (lower row).}  
	\label{intuition}
\end{figure}

Given the semantic object categories of interest, object detection aims at localizing each object instance from the input images. The prior research efforts mainly focus on the close-set setting, where the images containing the interested object categories are annotated and used to train a detector. The obtained detector only recognizes object categories that are annotated in the training set. In such a setting, more data needs to be collected and annotated if novel category\footnote{we treat \emph{category} and \emph{class} interchangeably in this paper} needs to be detected. However, data collection and annotation are very costly for object detection, which raises a significant challenge for traditional object detection methods.

\begin{figure*}[!t]
\centering
\includegraphics[width=0.99\linewidth]{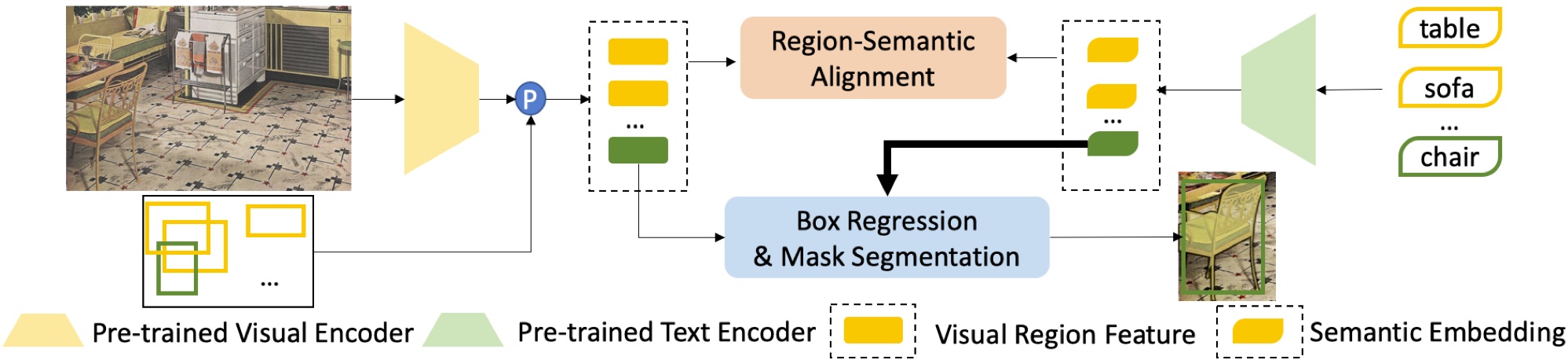}
\caption{\textbf{Overview of CondHead.} To detect objects of novel categories, we aim at \emph{conditionally parameterizing} the bounding box regression and mask segmentation based on the semantic embedding, which is strongly correlated with the visual feature and provides effective class-specific cues to refine the box and predict the mask.}
\label{fig:prompt_tuning}
\end{figure*}

To address the challenge, the open vocabulary object detection methods are widely explored recently, these methods~\cite{zareian2021open,gaoopen,minderer2022simple,du2022learning,zang2022open,zhong2022regionclip,gu2021open,huynh2022open,bansal2018zero,rahman2020improved} aim at generalizing object detection on novel categories by only training on a set of labeled categories. The core of these methods is transferring the strong image-text aligned features~\cite{radford2021learning,jia2021scaling} to classify objects of novel categories. To achieve bounding box regression and mask segmentation on novel categories, they simply employ the class-agnostic network heads. 
Although class agnostic heads can be readily applied to novel target object categories, they offer limited capacity to learn category-specific knowledge like object shape, and thus provide sub-optimal performance. On the other hand, training class-wise heads is not feasible as we do not have bounding box and mask annotation for the target categories. 
As shown in Figure~\ref{intuition}, our intuition is that the class-wise knowledge could naturally generalize across object categories, and may be leveraged to achieve much higher quality box regression and mask segmentation on the target categories, in a category-specific manner.
However, we find a brute force way of training class-wise heads on the base categories and manually gathering the class-specific prediction with closet semantic similarity during inference provides limited gain. The reason is that there still remains gap between the base and target categories, such as appearance and context.

Motivated by the strong semantic-visual aligned representation~\cite{zareian2021open,du2022learning,zhong2022regionclip,gu2021open} in open vocabulary object detection, we propose to exploit the semantic embedding as a conditional prior to parameterize class-wise bounding box regression and mask segmentation. Such conditioning is learned on base categories and easily generalizes to novel categories with the semantic embedding.
Our method, named \emph{CondHead}, is based on dynamic parameter generation of neural networks~\cite{jia2016dynamic,jaderberg2015spatial,dai2017deformable, yang2019condconv,chen2020dynamic}. To achieve strong efficiency, it exploits both large complex network head for their representative power and small light network head for their efficiency.
The complex head is employed by conditional weight aggregation over a set of static heads. The light head is employed by dynamically basing its parameters on the semantic embedding. The final prediction is obtained by combining the predictions from the two stream results. Through optimization on the base categories, the set of static heads are expected to learn sophisticated expert knowledge to cope with complex shapes and appearance, the dynamic head is endowed with general class-specific knowledge such as color and context.

Our CondHead is flexible regarding the choice of semantic-visual representation. The stronger quality of the aligned representation is expected to bring higher performance, as the conditional knowledge from the semantic embedding could generalize better to the target visual features. This is demonstrated by the clear grow of improvement over three baselines with increasing quality of pre-trained semantic-visual encoder networks, OVR-CNN~\cite{zareian2021open}, ViLD~\cite{gu2021open} and RegionCLIP~\cite{zhong2022regionclip}, on both COCO~\cite{lin2014microsoft} and LVIS~\cite{gupta2019} datasets. Remarkably, CondHead brings an average 2.8 improvement w.r.t both box and mask AP for the strong RegionCLIP baseline, with only about 1\% more computation. We also demonstrate intriguing qualitative results, showing how the semantic conditioning positively affects the regression and segmentation tasks.

Our contributions are three-fold. 1) To the best of our knowledge, we are the first to leverage semantic-visual aligned representation for open vocabulary box regression and mask segmentation. 2) We design a differentiable semantic-conditioned head design to efficiently bridge the strong category-specific prediction learned on base categories to the target novel categories. 3) We extensively validate the proposed method on various benchmark datasets and conduct thorough ablation and analysis to understand how the semantic conditioning helps detect and segment the novel categories.

\section{Related Work}

\myparagraphonept{Detecting Objects with Zero-shot and Open Vocabulary Setting} Despite the remarkable success of fully-supervised object detection~\cite{girshick2015fast,ren2015faster,cai2018cascade,liu2016ssd,redmon2016you,redmon2017yolo9000,carion2020end,zhu2020deformable}, the instance-level annotation for novel object categories is costly. Recent years, open vocabulary object detection emerges as an effective solution, which aims at detecting novel categories without corresponding annotation.
The representative methods employ region-based alignment, \eg, OVR-CNN~\cite{zareian2021open}, ViLD~\cite{gu2021open}, RegionCLIP~\cite{zhong2022regionclip} and DetPro~\cite{du2022learning}. 
They are based on the two-stage object detection framework~\cite{ren2015faster} and aim to match the object region feature with the generalizable semantic category embedding.
They mainly differ in the learning of semantic-visual representation. Specifically, OVR-CNN~\cite{zareian2021open} optimizes the semantic-visual grounding objective based multimodal transformer and applies the learned vision to language projection to facilitate novel category recognition. ViLD~\cite{gu2021open} transfers the stronger visual-semantic representation from large-scale vision-language pre-training~\cite{radford2021learning}. Based on ViLD, DetPro~\cite{du2022learning} introduces a learnable prompt mechanism~\cite{zhou2022learning} to enhance the semantic representation and achieve higher performance. RegionCLIP~\cite{zhong2022regionclip} instead focuses on more effective region-wise pre-training to obtain more discriminative representation. In addition to pursuing more stronger region feature alignment,
concurrent works~\cite{gaoopen,minderer2022simple,zang2022open,huynh2022open} explore other aspects of open vocabulary detection. For example, Gao \et~\cite{gaoopen} and Huynh \et~\cite{huynh2022open} develop pseudo-labeling approach to generate pseudo ground-truth labels for object detection and instance segmentation. Instead of labeling perspective, some works focus on transformer architecture design for open vocabulary detection. 
OV-DETR~\cite{zang2022open} proposes a binary matching objective to enable query matching with novel objects on the end-to-end DETR framework~DETR~\cite{carion2020end}.
Minderer \et~\cite{minderer2022simple} directly transfers pre-trained vision transformers to simplify open vocabulary detection. 

The prior works generally focus on the recognition of open vocabulary objects, while the localization or segmentation are typically conducted in a class-agnostic fashion. In this work, we explore this orthogonal perspective and aim at developing effective methods for more accurate box regression and mask segmentation.

\myparagraphonept{Dynamic Network Design}
Traditional neural network architectures employ static weights, \ie, the weights are independent of input signal and are fixed after training. Such static neural networks are limited at flexibility, in dealing with diverse input signals like images. Some previous works explore the idea of dynamic instantiation of networks to improve the flexibility. For example, Jia \et~\cite{jia2016dynamic} proposes to actively predict the convolution filters based on the input. In addition to explicit network parameter generation, Jaderberg \et~\cite{jaderberg2015spatial} introduces a dynamic parametric transformation (STN) to adaptively transform the image feature map, recovering affine distortions in the input and thus facilitating recognition. 
STN~\cite{jaderberg2015spatial} operates on global image space, the Deformable Convolution~\cite{dai2017deformable,zhu2019deformable} instead works in local scale. It learns spatial offsets to adjust the sampling locations of convolution kernels.
The dynamic designs like STN~\cite{jaderberg2015spatial} and Deformable Convolution~\cite{dai2017deformable,zhu2019deformable} introduce explicit control over the networks. Some recent works introduce implicit dynamics 
to achieve strong network capacity.
Notably,~\cite{yang2019condconv,chen2020dynamic} develop conditionally parameterized convolution filters to increase the network capacity. Specifically, the convolution kernel is instantiated by dynamically aggregating a set of base filters. 

Our work is related to the above dynamic network architecture designs, in that we also aim at achieving input-conditioned processing, to better adapt to the input data. Unlike prior works, we focus on the object instance-level design of dynamic processing. Moreover, instead of operating on the same visual signal, we instead introduce semantic signal for the dynamic network design.

\section{Methodology}

\subsection{Problem Definition}
Given the bounding box annotation on a set of base object categories $\mathbb{F} = \{C_1, C_2,... C_N\}$, open vocabulary object detection aims at training on the base data and generalizing to any open set of target object categories $\mathbb{F}^*=\{C_1, C_2,... C_M\}$. Open vocabulary instance segmentation is also expected
if the mask annotation is given.

The state-of-the-art two-stage methods~\cite{zareian2021open,du2022learning,zhong2022regionclip,gu2021open} can be simplified as a region-semantic alignment framework.
The model is initialized with an image encoder ($\mathcal{T}_i$) and a language encoder ($\mathcal{T}_l$), which extract image feature and semantic embedding: $\textbf{F}=\mathcal{T}_i(I)$, $\textbf{s}_n=\mathcal{T}_l(C_n)$, here $I$ denotes the input image. These encoders~\cite{radford2021learning,jia2021scaling} are typically pre-trained on large-scale image-caption~\cite{chen2015microsoft,sharma2018conceptual,radford2021learning} data and thus extract strongly correlated visual-semantic representations.
With region proposal network~\cite{ren2015faster} trained on the base categories to extract object proposals, the open-vocabulary detection is trained in a region-wise manner. 

Concretely, given an object proposal bounding box $\textbf{p}=(x_1, y_1, x_2, y_2)$, feature pooling is conducted on the image feature map $\textbf{F}$ to obtain the region feature $\textbf{f}$.
The object proposal box is then matched with the ground-truth object instance set $\aleph$ w.r.t IOU (Intersection Over Union) metric:
\begin{align}
(\textbf{b}, c)=\argmax_{(\textbf{b}', c') \in \aleph }\mathrm{IOU}(\textbf{p}, \textbf{b}')
\end{align}
where $\textbf{b}=(x^*_1, y^*_1, x^*_2, y^*_2)$ means the obtained ground-truth bounding box and $c$ is corresponding object category. 
Then, optimization is conducted to further minimize the alignment between ground-truth semantic embedding $\textbf{s}_c$ and region feature $\textbf{f}$, \eg, through similarity-based cross entropy loss~\cite{gu2021open}.

To generalize the bounding box regression, prior works employ class-agnostic network head $\mathcal{B}$ with parameter $\theta$ and optimize it on the base category data:
\begin{align}
\min_{\theta} \mathcal{L}_{\mathcal{B}}(\mathcal{B}_\theta, \mathcal{X}) =  \mathcal{L}_{\mathcal{B}}(\mathcal{B}_\theta(\boldsymbol{p, f}), \boldsymbol{b}), 
\end{align}
where $\mathcal{X}=(\textbf{p},\textbf{f},\textbf{b})$ denotes the tuple data of proposal, region feature and bounding box. $\mathcal{L}_{\mathcal{B}}$ is typically defined as \textbf{L1} or \textbf{L2} error between the regressed bounding box and ground-truth box.

To learn instance segmentation, the mask region feature $\textbf{v}$ is pooled with the refined bounding box from the box head, Then the class-agnostic mask segmentation network head $\mathcal{M}$ with parameter $\vartheta$ is trained by:
\begin{align}
\min_{\vartheta} \mathcal{L}_{\mathcal{M}}(\mathcal{M}_\vartheta, \mathcal{Y}) =  \mathcal{L}_{\mathcal{M}}(\mathcal{M}_\vartheta(\boldsymbol{v}), \boldsymbol{m}), 
\end{align}
where $\textbf{m}$ denotes the corresponding ground-truth mask segmentation and $\mathcal{Y}=(\textbf{v},\textbf{m})$ denotes the paired region feature and mask. $\mathcal{L}_{\mathcal{M}}$ is typically defined as pixel-wise segmentation error between the predicted mask and ground-truth.

Although the class-agnostic bounding box regression and mask segmentation heads readily generalize to the target novel object categories, they offer sub-optimal results due to the less representative capability. 
Since the semantic embeddings are strongly correlated with the visual features, we propose to achieve \emph{generalizable class-wise prediction} by conditioning the network parameters on the base category semantic embeddings during training:

\begin{align}
\min_{\alpha} \mathcal{L}_{\mathcal{B}}(\mathcal{B}_{\theta(\textbf{s}_c)}, \mathcal{X})
\label{eq4}
\end{align}
\begin{align}
\min_{\beta} \mathcal{L}_{\mathcal{M}}(\mathcal{M}_{\vartheta(\textbf{s}_c)}, \mathcal{Y}) 
\label{eq5}
\end{align}
Where $\alpha$ and $\beta$ are parameters of the conditioning function $\theta(\cdot)$ and $\vartheta(\cdot)$. 
Through this approach, the model learns to generalize regression and segmentation to novel categories in a class-wise manner, by predicting corresponding parameters $\theta(\textbf{s}_m)$ and $\vartheta(\textbf{s}_m)$ during inference. 



\subsection{CondHead Formulation}
Our proposed CondHead aims to bridge the gap between base and target categories, with the semantic conditioning mechanism.
We instantiate the conditioning function with neural networks to achieve flexible optimization and inference.
To construct such conditioning framework, we observe the challenges in network capacity: 1) More complex head architecture (\eg, head with several hidden layers) offers stronger representative ability, but it requires large number of parameters, which are not efficient to generate and hard to optimize. For example, a hidden fully layer with input and output dimension of $256$ involves more than $60$k scalar weights.
2) Less complex head architecture (\eg, head with a single layer) is easier to optimize and can be efficiently generated, but provides limited capacity.

To address the challenge, we design a \emph{dual conditioning framework}, to leverage both the complex heads and light heads. The framework is composed of dynamically aggregated head and dynamically generated head.

\myparagraphonept{Dynamically Aggregated Head} To leverage the large network capacity of complex heads, we propose to generate dynamic weights to aggregate a set of complex heads $\{\mathcal{B}_1, {\mathcal{B}_2},...,\mathcal{B}_H\}$, with associated parameters $\{\theta_1, {\theta_2},...,\theta_H\}$. These heads act as experts that are good at refining bounding box and predicting mask segmentation for objects categories with certain shapes or appearance.
Specifically, the aggregation weights is generated by:
\begin{align}
\textbf{w} = \mathcal{A}_\phi(\textbf{s}_c)
\end{align}
where $\textbf{w}\in\mathbb{R}^{H}$, and $\mathcal{A}$ is a small neural network with parameter $\phi$.
To regularize the weight aggregation space as a convex hull for optimization, the weight $\textbf{w}$ is then normalized with Softmax function:
\begin{align}
w_{h} =\frac{e^{w_{h}}}{\sum_{h^{'}=1}^{H}e^{w_{h^{'}}}}
\end{align}
which ensures the weights sums to $1$ and each element is larger than $0$.
With the normalized aggregation weight, the expert heads are then combined to a single head $\hat{\mathcal{B}}$ with parameter $\hat{\theta}$:
\begin{align}
\hat{\theta} =\sum_{h=1}^{H}w_h*\theta_h
\end{align}
Similarly, for a set of mask heads $\{\mathcal{M}_1, {\mathcal{M}_2},...,\mathcal{M}_H\}$, with associated parameters $\{\vartheta_1, {\vartheta_2},...,\vartheta_H\}$, the weight generation and aggregation can be conducted to obtain a single mask head $\hat{\mathcal{M}}$ with parameter $\hat{\vartheta}$.
When the dynamic aggregation is applied to both box regression and mask segmentation, separate weight generation networks are used, as box regression and mask segmentation may not share the same attention across the expert models.

\myparagraphonept{Dynamically Generated Head} To introduce stronger conditioning on the semantic embedding, we propose to directly generate the network parameter. With bounding box regression head as an example, the parameter is first generated as:
\begin{align}
\dot{\theta} = \mathcal{D}_\varphi(\textbf{s}_c)
\end{align}
the parameter $\dot{\theta}$ is used to instantiate a new box head $\dot{\mathcal{B}}$. Since $\dot{\theta}$ is directly generated from the semantic embedding, it easily encodes general class-specific information like aspect ratio and color. Similarly, a dynamic mask segmentation head $\dot{\mathcal{M}}$ can be obtained with generated parameter $\dot{\vartheta}$.

\myparagraphonept{Combining the Dynamic Predictions} We then combine the above dynamically aggregated heads $\hat{\mathcal{B}}$, $\hat{\mathcal{M}}$ and dynamically generated heads $\dot{\mathcal{B}}$, $\dot{\mathcal{M}}$ to obtain the final result. We consider simple weighted averaging on their prediction:

\begin{align}
\overline{\mathcal{B}}_{\theta(\textbf{s}_c)}(\cdot) = \lambda*\hat{\mathcal{B}}(\cdot) + (1-\lambda)*\dot{\mathcal{B}}(\cdot)
\end{align}
\begin{align}
\overline{\mathcal{M}}_{\vartheta(\textbf{s}_c)}(\cdot) = \mu*\hat{\mathcal{M}}(\cdot) + (1-\mu)*\dot{\mathcal{M}}(\cdot)
\end{align}
$\overline{\mathcal{B}}$ and $\overline{\mathcal{M}}$ are for the box regression and mask segmentation head respectively. $\lambda$ and $\mu$ are hyper-parameters. 

After optimization on the base category data with equation~\ref{eq4} and equation~\ref{eq5}, $\overline{\mathcal{B}}$ and $\overline{\mathcal{M}}$ are employed during inference on the target categories by conditioning on the target semantic embedding as $\overline{\mathcal{B}}_{\theta(\textbf{s}_m)}(\cdot)$ and $\overline{\mathcal{M}}_{\vartheta(\textbf{s}_m)}(\cdot)$

\myparagraphonept{Optimizing the Expert Heads with Temperature Annealing} During the early optimization stage, the expert heads within the dynamically aggregated head all require training signals to initiate the learning of basic regression and segmentation capability. While in the later optimization stage, the experts are expected to provide specialized capability in dealing with different objects.
We thus facilitate the optimization of the expert heads by a temperature annealing Softmax strategy. Concretely, the normalization of aggregation weights is performed with temperature $\tau$
\begin{align}
w_{h} =\frac{e^{w_{h}/\tau}}{\sum_{h^{'}=1}^{H}e^{w_{h^{'}}/\tau}}
\end{align}
During the early training stages, the temperature is set as a large value and with the progress of training to provide nearly uniform gradients for all the expert heads, the temperature is gradually annealed to a smaller value to achieve the desired specialized learning.

\subsection{Relation to Close Works}
\label{sec:relation_to_close_works}

\myparagraphonept{Conditionally Parameterized Convolution} CondHead is closely related to Dynamic Convolution~\cite{chen2020dynamic} and CondConv~\cite{yang2019condconv}. They differ from CondHead in motivation and application. They aim to improve the representation capacity of convolution kernels through conditional parameterization, the conditioning is performed on the visual features of each network layer. Moreover, they are applied to the image recognition task. 
While CondHead aims to bridge the perception gap between the base training object categories and target novel object categories by conditioning on the strongly visual-aligned semantic embedding. CondHead is applied to the more challenging bounding box regression and mask segmentation task, which is not explored before.

\myparagraphonept{Partially Supervised Instance Segmentation} Recent works~\cite{hu2018learning,kuo2019shapemask} explore partially supervised instance segmentation, where the base categories have both box and mask annotation while the novel categories have only box annotation. Mask$^X$ R-CNN~\cite{hu2018learning} propose a parameterized transformation function to transfer the box regression weights to mask segmentation weights. Shapemask~\cite{kuo2019shapemask} explores strong shape priors to achieve better class-agnostic object segmentation, which generalizes better to novel categories without mask annotation. Our focused open vocabulary object detection and instance segmentation are much more challenging than the partially supervised instance segmentation, which is based on good detection quality and strong object region representation.
Mask$^X$ R-CNN~\cite{hu2018learning} cannot be applied here as box annotation is not available. Shapemask~\cite{kuo2019shapemask} could be applied here, we compare to it for instance segmentation and evaluate an augmented version of CondHead with Shapemask.

\section{Experiments}
We study four questions in experiments. 1) Is CondHead able to improve the performance of open vocabulary object detection? is it efficient? 2) How does the quality of semantic-visual representation affects the performance? 3) How does CondHead improves the box regression and mask segmentation? 4) How does each component and hyper-parameter take effect? 
Throughout the experiments, unless otherwise stated, we adopt the architectural instantiation shown in Table~\ref{condhead_arch}.


\begin{table}[]
\centering
\setlength{\tabcolsep}{2pt}
\begin{tabular}{l|cccccc}
\toprule
Head & $\mathcal{A}$ & $\mathcal{D}$ & $\{\mathcal{B}_h\}$ & $\{\mathcal{M}_h\}$ & $\dot{\mathcal{B}}$ & $\dot{\mathcal{M}}$  \\
Architecture & 2fc & 2fc & 2fc & 3conv & 1fc & 1conv\\
\bottomrule
\end{tabular}
\caption{Architectural instantiate of CondHead. fc and conv denote fully connected and convolutional networks, respectively, with a hidden dimension of 256. The digit means the number of layers.}
\label{condhead_arch}
\end{table}

\begin{table*}[t!]
\centering
\begin{tabular}{l|c>{\color{gray}}c>{\color{gray}}c|c>{\color{gray}}c>{\color{gray}}c}
\toprule
 & \multicolumn{3}{c|}{Object Detection} & \multicolumn{3}{c}{Instance Segmentation} \\  
        Method     & Novel     & Base     & All     & Novel     & Base     & All     \\ \hline
         DELO~\cite{zhu2020don} (CVPR20)    &  3.41    &   13.8    &   13.0     & -      &   -    &    -        \\
         \rowcolor{grayLight}
         PL~\cite{rahman2020improved} (AAAI20)  &   4.12 &   35.9    &  27.9    & -    &   -    &   -    \\ \hline
            OVR-CNN~\cite{zareian2021open} (CVPR21)   &   22.8 &    46.0  &   39.9    &   -    &   -    &     -    \\
            \rowcolor{grayLight}
            OVR-CNN*  &   22.6 &    45.9  &   39.8    &   20.1    &   42.3    &  36.5         \\
        \rowcolor{grayDark}
          CondHead (Ours)    &   \TableEntry{24.0}{1.4}    &  \TableEntryGray{46.5}{0.6}    &   \TableEntryGray{40.6}{0.8}   &   \TableEntry{21.4}{1.3}  & \TableEntryGray{42.9}{0.6}     &  \TableEntryGray{37.3}{0.8}   \\ \hline
          ViLD~\cite{jia2021scaling}  (ICLR22) &   27.6 &    59.5  &   51.3    &   -    &   -    &     -    \\
          \rowcolor{grayLight}
          ViLD*  &   27.7 &    59.7  &   51.4    &   24.1    &   56.7    &    48.2    \\
          \rowcolor{grayDark}
          CondHead (Ours)  &   \TableEntry{29.8}{2.1}    &  \TableEntryGray{60.8}{1.1}    &   \TableEntryGray{52.7}{1.3}   &   \TableEntry{25.9}{1.8}  & \TableEntryGray{57.9}{1.2}     &  \TableEntryGray{49.5}{1.3}   \\ \hline
          RegionCLIP~\cite{zhong2022regionclip} (CVPR22) &   31.4 &    57.1  &   50.4    &   -    &   -    &     -    \\
          \rowcolor{grayLight}
          RegionCLIP* &   31.3 &    56.5  &   49.9    &   27.5    &   54.1    &     47.1    \\
          \rowcolor{grayDark}
          CondHead (Ours)  &   \TableEntry{33.7}{2.4}    &  \TableEntryGray{58.0}{1.5}    &   \TableEntryGray{51.7}{1.8}   &   \TableEntry{29.7}{2.2}  & \TableEntryGray{55.8}{1.7}     &  \TableEntryGray{49.0}{1.9}   \\ \bottomrule
\end{tabular}
\caption{\textbf{Open vocabulary object detection results on COCO.} * denotes our re-evaluation by adding the class-agnostic mask head.}
\label{table:main:coco}
\end{table*}

\subsection{Datasets and Setup}
We adopt object detection benchmark datasets COCO~\cite{lin2014microsoft} and LVIS~\cite{gupta2019} to evaluate our method. Following prior works~\cite{zareian2021open,gu2021open}, we manually split the datasets into base and target categories\footnote{we use the pre-processed data from~\cite{zareian2021open} and~\cite{zhong2022regionclip}.}.
We also evaluate the generalization through inference on two other common object benchmarks, PASCAL VOC~\cite{everingham2010pascal} and  Objects365~\cite{shao2019objects365}. 

\begin{table*}[]
\centering
\setlength{\tabcolsep}{1pt}
\begin{tabular}{l|c|c>{\color{gray}}c>{\color{gray}}c>{\color{gray}}c|c>{\color{gray}}c>{\color{gray}}c>{\color{gray}}c}
\toprule
 & & \multicolumn{4}{c|}{Object Detection} & \multicolumn{4}{c}{Instance Segmentation} \\
Method &  Backbone & AP$_r$   & AP$_c$   & AP$_f$   & AP   & AP$_r$   & AP$_c$   & AP$_f$   & AP   \\ \hline
Supervised & \multirow{2}{*}{RN50-FPN} &  3.3   &  22.5    &   34.5  &  23.3   &  4.2   &  22.8   &  32.7   & 21.8    \\ 
Supervised +RFS~\cite{gupta2019} &  &  11.6   &  23.5   &  32.5     & 24.3     &  12.8   &  23.3   & 31.0     &  23.1   \\ \hline
ViLD~\cite{gu2021open} & \multirow{3}{*}{RN50-FPN} &  16.7   &  26.5    &   34.2    &  27.8   &  16.6    &  24.6   &   30.3  &  25.5     \\
\rowcolor{grayLight}
ViLD* & RN50-FPN &  16.6   &  27.0   &  33.0 &    27.5 &  16.9   &    25.0 &  29.1   &    25.2 \\
\rowcolor{grayDark}
CondHead (Ours) &  &  \TableEntry{18.8}{2.2}   & \TableEntryGray{28.3}{1.3}    &  \TableEntryGray{33.7}{0.7}   &  \TableEntryGray{28.8}{1.3}   &  \TableEntry{19.1}{2.2}   &  \TableEntryGray{26.2}{1.1}   &  \TableEntryGray{29.9}{0.8}   & \TableEntryGray{26.4}{1.2}    \\ \hline
 RegionCLIP~\cite{zhong2022regionclip} & \multirow{3}{*}{RN50-C4} & 17.1 &  27.4    &  34.0   & 28.2  & - & - &  -  & -    \\
 \rowcolor{grayLight}
 RegionCLIP* & RN50-C4 &   17.0  & 27.2 &  34.3    &  28.2   &    17.4 &  26.0    &  31.6   & 26.7 \\
 \rowcolor{grayDark}
 CondHead (Ours) & & \TableEntry{19.9}{2.9} &  \TableEntryGray{28.6}{1.4}    &  \TableEntryGray{35.2}{0.9}  & \TableEntryGray{29.7}{1.5} & \TableEntry{20.0}{2.6} & \TableEntryGray{27.3}{1.3} &  \TableEntryGray{32.2}{0.6}  &  \TableEntryGray{27.9}{1.2}   \\ \hline
 ViLD~\cite{gu2021open} & \multirow{3}{*}{RN152-FPN} & 19.8 & 27.1 & 34.5 & 28.7 & - & - & - & - \\
\rowcolor{grayLight}
 ViLD* & RN152-FPN & 20.0 & 27.0 & 34.0 & 28.5 & 19.8 & 25.1 & 32.1 & 26.9 \\
\rowcolor{grayDark}
 CondHead (Ours) & & \TableEntry{21.9}{1.9} & \TableEntryGray{28.4}{1.4} & \TableEntryGray{34.6}{0.6} & \TableEntryGray{29.7}{1.2} & \TableEntry{21.6}{1.8} & \TableEntryGray{26.2}{1.1} & \TableEntryGray{33.0}{0.9} & \TableEntryGray{28.1}{1.2} \\ \hline
RegionCLIP~\cite{zhong2022regionclip} & \multirow{3}{*}{RN50x4-C4} & 22.0 & 32.1 & 36.9 & 32.3 & - & - & - & - \\ 
\rowcolor{grayLight}
RegionCLIP* & RN50x4-C4 & 22.1 & 31.8 & 37.0 & 32.2 & 21.8 & 30.2 & 35.1 & 30.7 \\
\rowcolor{grayDark}
CondHead (Ours) & & \TableEntry{25.1}{3.0} & \TableEntryGray{33.4}{1.6} & \TableEntryGray{37.8}{0.8} & \TableEntryGray{33.7}{1.5} & \TableEntry{24.4}{2.6} & \TableEntryGray{31.6}{1.4} & \TableEntryGray{35.9}{0.8} & \TableEntryGray{32.0}{1.3} \\ \bottomrule
\end{tabular}
\caption{\textbf{Open vocabulary object detection result on LVIS.} * denotes our re-evaluation by adding the mask head. Supervised and Supervised + RFS are baselines that have access to the training data of target categories. RFS means the repeat factor sampling method~\cite{gupta2019}.}
\label{table:main:lvis}
\end{table*}

\myparagraphonept{Setup} Since the prior works mainly focus on the classification of novel object categories, we evaluate CondHead by adopting the prior state-of-the-art open vocabulary methods and treating them as baselines. We employ the following methods with increasing quality of semantic-visual alignment:
\begin{itemize}
    \item \myparagraph{OVR-CNN}~\cite{zareian2021open} proposes to pre-train the multi-modal encoder with image-caption pair data. The pre-trained rich visual-semantic representation is then transferred to recognize the objects with open vocabulary.
    \item \myparagraph{ViLD}~\cite{gu2021open} achieves much higher performance than OVR-CNN by transferring the stronger visual-semantic representation of CLIP~\cite{radford2021learning}, which was pre-trained on a much larger scale of image-caption data.
    \item \myparagraph{RegionCLIP}~\cite{zhong2022regionclip} further improves the alignment between the semantic and visual embeddings by conducting region-wise pre-training. The improved representation further improves the recognition of novel objects and thus brings significant performance gain.
\end{itemize}

\subsection{Results}
\myparagraph{Results on COCO} We re-evaluate the baseline methods by adding class-agnostic mask segmentation heads. Then we replace the bounding box regression and mask segmentation heads with the proposed CondHead and re-run the experiments.
As shown in Table~\ref{table:main:coco}, our method surpasses all baseline methods, especially on the novel object category set. We observe the improvements grow with the three baselines, \eg, the object detection AP improvements are 1.4 with OVR-CNN, 2.1 with ViLD, and 2.4 with RegionCLIP. A similar observation holds for the instance mask segmentation task. This phenomenon verifies that CondHead can leverage semantic embeddings that better align with visual representations to achieve stronger bounding box regression and mask segmentation ability. 

\begin{figure*}[t!]
\centering
\includegraphics[width=0.95\linewidth]{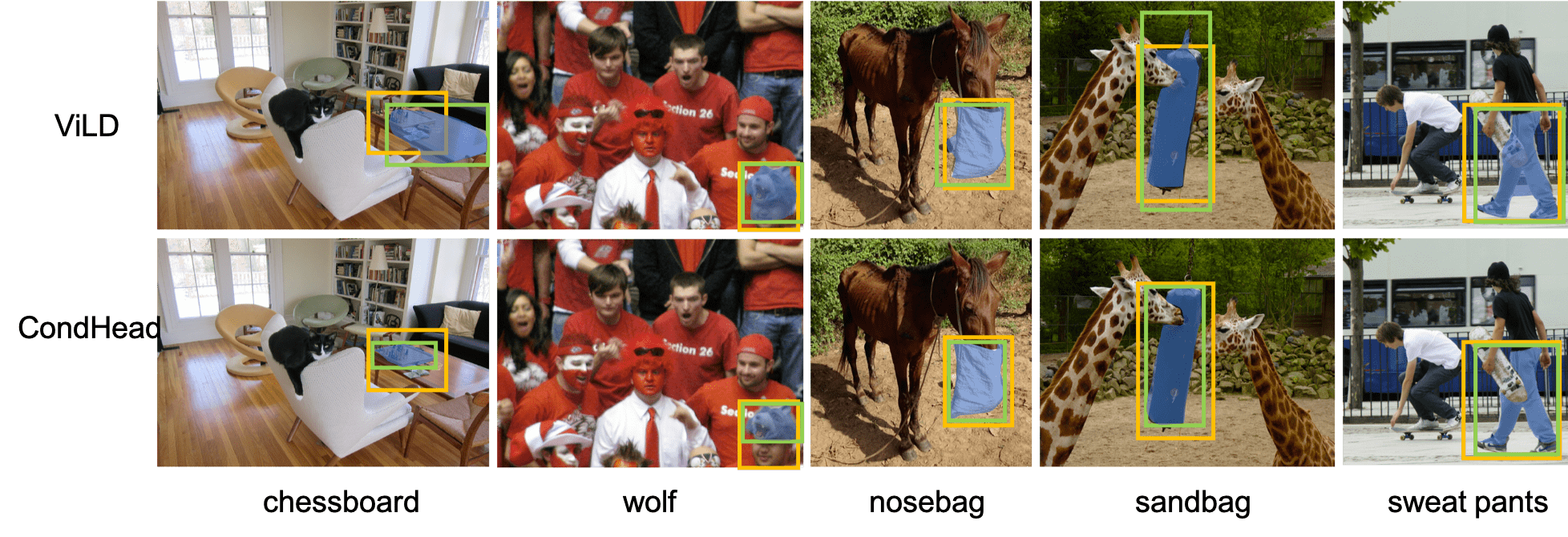}
\caption{Qualitative comparison with baseline ViLD~\cite{gu2021open}. The bounding box regression and mask segmentation results are overlaid on the images (Yellow: Proposals. Green: Regressed bounding box. Blue: segmentation mask). Best viewed with zoom-in.}
\vspace{-5pt}
\label{fig:qualitative}
\end{figure*}

\begin{table*}[]
\centering
\setlength{\tabcolsep}{1.5pt}
\begin{tabular}{l|cc|ccc|ccc}
\toprule
 & \multicolumn{2}{c|}{PASCAL VOC} & \multicolumn{3}{c|}{COCO} & \multicolumn{3}{c}{Objects365} \\
Method &     AP$_{50}$      &     AP$_{75}$     &   AP    &   AP$_{50}$    &   AP$_{75}$   &     AP    &   AP$_{50}$    &   AP$_{75}$  \\ \hline
Supervised &    78.5       &    49.0      &    46.5   &    67.6    &  50.9     &    25.6   &   38.6    &   28.0    \\
\rowcolor{grayLight}
ViLD~\cite{gu2021open} &     72.2     &    56.7        &    36.6  &     55.6    &   39.8    &   11.8   &  18.2   &   12.6   \\
\rowcolor{grayDark}
CondHead (Ours) &    \TableEntry{74.6}{2.4}       &    \TableEntry{58.5}{1.8}    & \TableEntry{39.1}{2.5}      &   \TableEntry{59.1}{3.5}    &  \TableEntry{42.2}{2.4}    &    \TableEntry{13.2}{1.4}   &   \TableEntry{20.4}{2.2}    &    \TableEntry{14.2}{1.6}\\ \bottomrule
\end{tabular}
\caption{Cross-dataset evaluation results. The model is trained on LVIS and directly evaluated on the target datasets.}\vspace{-5pt}
\label{table:main:transfer}
\end{table*}

\myparagraphonept{Results on LVIS} Following the evaluation on COCO, We then evaluate CondHead on LVIS dataset. As shown in Table~\ref{table:main:lvis}, our method brings about 2.0-3.0 absolute improvement on the target novel object categories (\ie, AP$_r$), for both bounding box detection and instance segmentation. The improvement holds for stronger backbone networks. We also observe a similar growth of improvement as COCO, \ie, the RegionCLIP method gains more improvement with CondHead than that of ViLD. This is likely because RegionCLIP provides stronger semantic-visual alignment with its region-wise pre-training, and thus CondHead better generalizes to the target categories.

\myparagraphonept{Cross Dataset Evaluation} We further evaluate CondHead with the cross dataset scenarios, where the open vocabulary object detector is expected to generalize on datasets that are different from the one used for training. We simply replace the vocabulary of a trained model with target dataset vocabulary to evaluate the performance. As shown in Table~\ref{table:main:transfer}, we observe the improvement of CondHead transfers well. For example, 3.5, 2.4, and 2.2 absolute improvements in AP$_50$ for COCO, PASCAL VOC, and Objects365 respectively.

\myparagraphonept{Qualitative Results}
To demonstrate how $CondHead$ improves bounding box regression and mask segmentation, we visualize the example qualitative results and compare that with the baseline method. As shown in Figure~\ref{fig:qualitative}, CondHead is better at refining the proposal box and segmenting the target objects. For example, the baseline ViLD wrongly regresses the target chessboard to the table, while CondHead correctly predicts the chessboard (Figure~\ref{fig:qualitative} first column). In addition, CondHead better segments the challenging sweat pants while ViLD predicts inferior segmentation mask (Figure~\ref{fig:qualitative} last column).

\begin{figure*}[!t]
\centering
\includegraphics[width=0.95\linewidth]{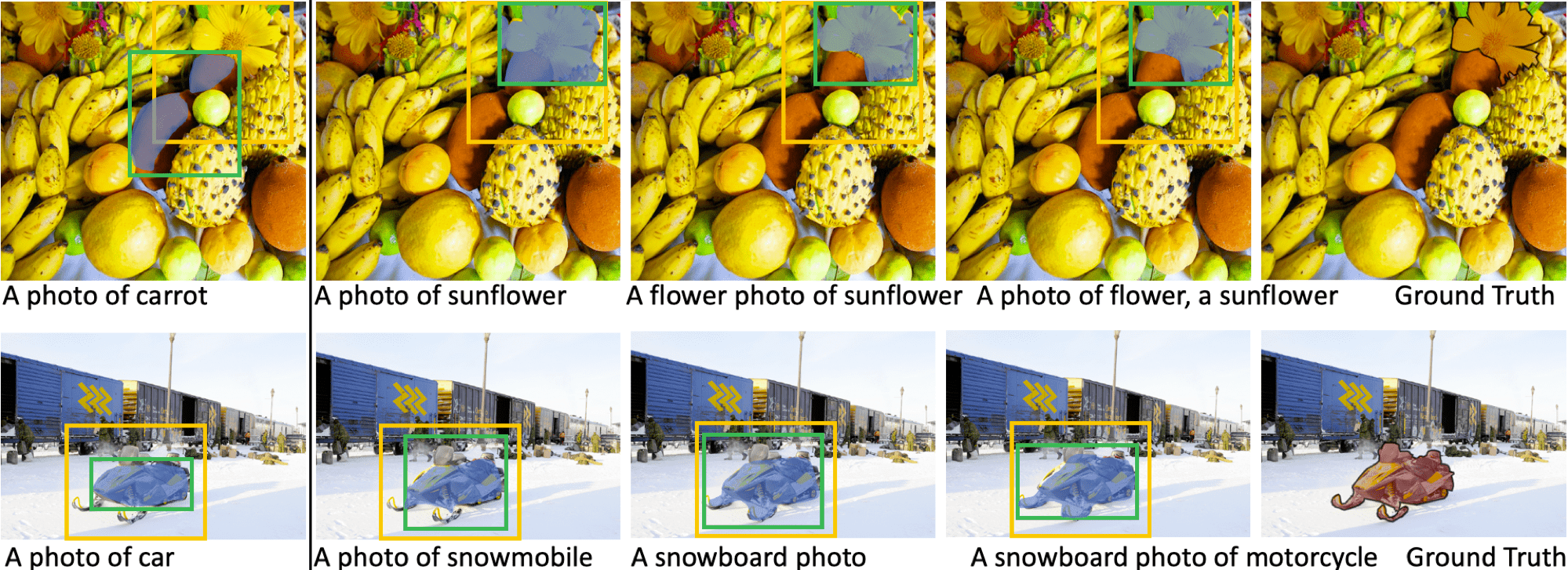}
\caption{Effect of tuning language descriptions. We select some intriguing examples for which tuning the input language descriptions could deteriorate or help object detection. Yellow: Proposals. Green: Regressed bounding box.}\vspace{-10pt}

\label{fig:description_tuning}
\end{figure*}

\subsection{Analysis}

\myparagraphonept{The Effects of Language Descriptions} We examine how CondHead is affected by language descriptions. As shown in Figure~\ref{fig:description_tuning}, when the input text is for a wrong category label, the box regression deteriorates, \eg, the proposal for sunflower is misguided to predict the carrot. On the other hand, it is interesting that when supplied with manually tuned descriptions, the detection and segmentation quality is improved, \eg, when employing snowboard and motorcycle as descriptions, the snowmobile is predicted better. This intriguing observation suggests CondHead learns strong category-conditioned prediction and a better performance could be achieved by simply tuning the semantic descriptions during inference.

\begin{table}[h]
\centering
\begin{tabular}{l|cc}
\toprule
 & AP$^b_r$ & AP$^m_r$ \\ \hline
ViLD* & 16.6 & 16.9 \\
\rowcolor{grayLight}
ClassWise & \TableEntryGray{17.1}{0.5} & \TableEntryGray{17.3}{0.4} \\
\rowcolor{grayDark}
CondHead & \TableEntry{18.8}{2.2} & \TableEntry{19.1}{2.2} \\ \hline
RegionCLIP* & 17.0 & 17.4 \\
\rowcolor{grayLight}
ClassWise & \TableEntryGray{17.3}{0.3} & \TableEntryGray{17.9}{0.5} \\
\rowcolor{grayDark}
CondHead & \TableEntry{17.9}{2.9} & \TableEntry{20.0}{2.6}\\
\bottomrule
\end{tabular}
\caption{The results of naive class-wise transfer experiment (discussed in Section~\ref{sec:intro}) on LVIS. (AP$^b_r$: box AP, AP$^m_r$: mask AP).}
\label{table:ablation:clswisebasline}
\end{table}

\myparagraph{Naive Class-wise Transfer} We also establish another intuitive experiment as discussed in Section~\ref{sec:intro} to demonstrate the effectiveness of the proposed method. Since our intuition (Figure~\ref{intuition}) is that the class-specific knowledge could be shared across categories, we train the class-wise heads on the base categories and then perform inference on the target categories in a semantically guided way. This is achieved by gathering the class-specific predictions with semantic similarity to the base categories. We use cosine similarity on the semantic embedding and select the category with the highest similarity. As shown in Table~\ref{table:ablation:clswisebasline}, we train the ClassWise baseline on both ViLD and RegionCLIP. ClassWise method provides a small improvement to class-agnostic heads compared to CondHead, demonstrating that CondHead better generalizes to the target categories.

\begin{table}[h]
\centering
\setlength{\tabcolsep}{2pt}
\begin{tabular}{l|cccc|cccc}
\toprule
 & V & SH & CH & CH$_S$ & R & SH & CH & CH$_S$ \\ \hline
\rowcolor{grayLight}
 AP$^m_r$ & 16.9 & 19.6 & 19.1 & 20.4 & 17.4 & 20.3 & 20.0 & 21.0 \\
\rowcolor{grayDark}
 $\Delta$ & - & 2.7 & 2.2 & \textbf{3.5} & - & 2.9 & 2.6 & \textbf{3.6}  \\
\bottomrule
\end{tabular}
\caption{Results with Shapemask. V and R denote ViLD and RegionCLIP, SH, CH, and CH$_S$ denote Shapemask head, CondHead and CondHead augmented with Shapemask, respectively.}
\label{table:ablation:shapemask}
\end{table}

\myparagraphonept{Augmenting CondHead with Shape Priors} As discussed in Section~\ref{sec:relation_to_close_works}, we compare to Shapemask~\cite{kuo2019shapemask} which introduces explicit class-agnostic shape priors.
We then explore an augmented version of CondHead by integrating the Shapemask, the semantic embedding is utilized to estimate the initial shape prior, then the dynamic expert aggregation is conducted on the latter coarse mask estimation and refinement. Please refer to the supplementary for more implementation details.
As shown in Table~\ref{table:ablation:shapemask}, Shapemask brings comparable improvement to CondHead, this is attributed to its strong prior and complex multi-stage mask refinement capability. Further integrating Shapemask into CondHead offers larger gain, \eg, 3.5 and 3.6 AP improvement for the two baselines. This implies CondHead can be improved with explicit shape priors.

\begin{table}[]
\centering
\begin{tabular}{lcccc}
\toprule
CondBox  & CondMask &  AP$^b_r$ &  AP$^m_r$ & FLOPs \\\hline
  &  & 17.0 & 17.4 & 193.5 \\
  \rowcolor{grayLight}
   \checkmark & & 19.6 & 17.8 & 194.1 \\
   \rowcolor{grayDark}
   \checkmark  & \checkmark  & 19.9 & 20.0& 195.8 \\
   \bottomrule
\end{tabular}
\caption{Component analysis. CondBox and CondMask mean applying CondHead on box and mask heads, respectively.}
\label{tab:component_analysis}
\end{table}

\begin{figure}[h]
  \centering
  \includegraphics[width=0.9\linewidth]{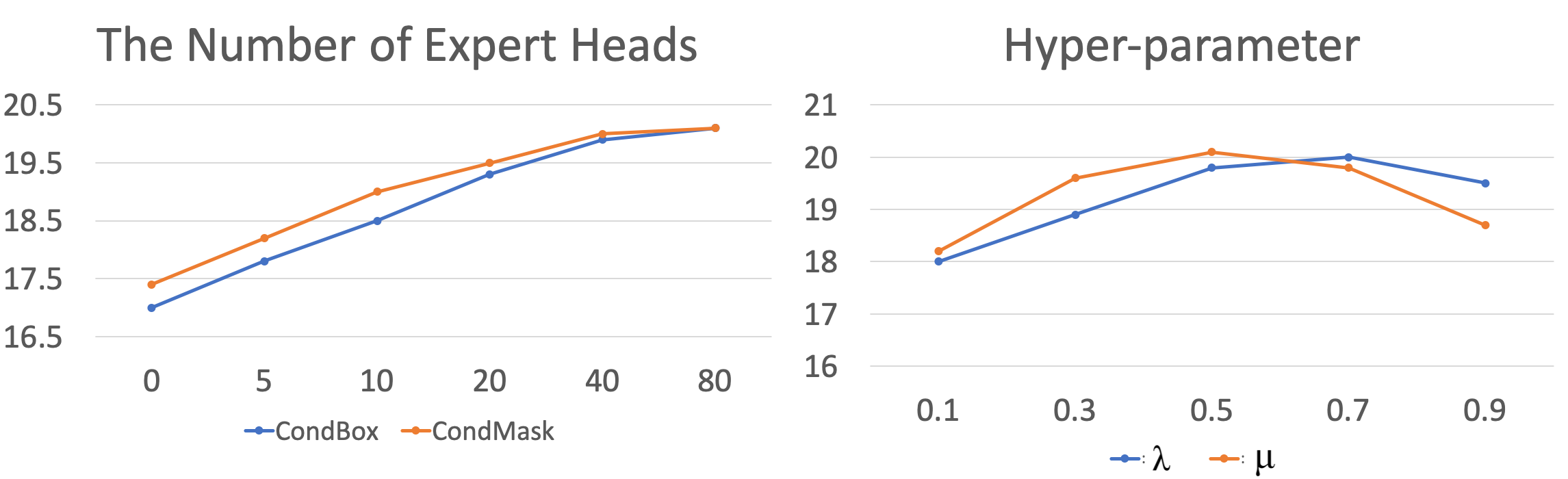}
  \caption{Component analysis. Effect of expert number, $\lambda$ and $\mu$.}\vspace{-5pt}
\label{fig:component_analysis}
\end{figure}

\myparagraphonept{Component Analysis}
As shown in Table~\ref{tab:component_analysis}, we perform an ablation study by applying CondHead on the box and mask heads independently, we find they improve box AP and mask AP independently.
We also observe CondHead introduces a very small computation overhead. To understand how the number of expert heads and aggregation hyper-parameter affects CondHead, a detailed analysis is conducted as shown in Figure~\ref{fig:component_analysis}. We find the performance gain is diminishing beyond 40 expert heads and the hyper-parameters $\lambda$ and $\mu$ work best at between 0.5 and 0.7. More analysis experiments are presented in the supplementary.

\section{Conclusion}
We introduce a conditionally parameterized neural network design to improve open vocabulary bounding box regression and mask segmentation, which is not explored before. Specifically, we leverage the pre-trained semantic embedding to guide the parameterization of the box and mask heads. The semantic embedding is strongly aligned with visual representation and thus provides effective cues for refining the bounding box and segmenting the objects. Our method named \emph{CondHead} is extensively validated on different datasets and setups. We hope our findings provide insights for future work on open vocabulary detection.

\nocite{wang2022sodar,wang2019distilling,wang2019few,wang2020devil}

{\small
\bibliographystyle{ieee_fullname}
\bibliography{egbib}
}

 \begin{figure*}[]
  \centering
  \subfloat[]{\includegraphics[width=0.99\linewidth]{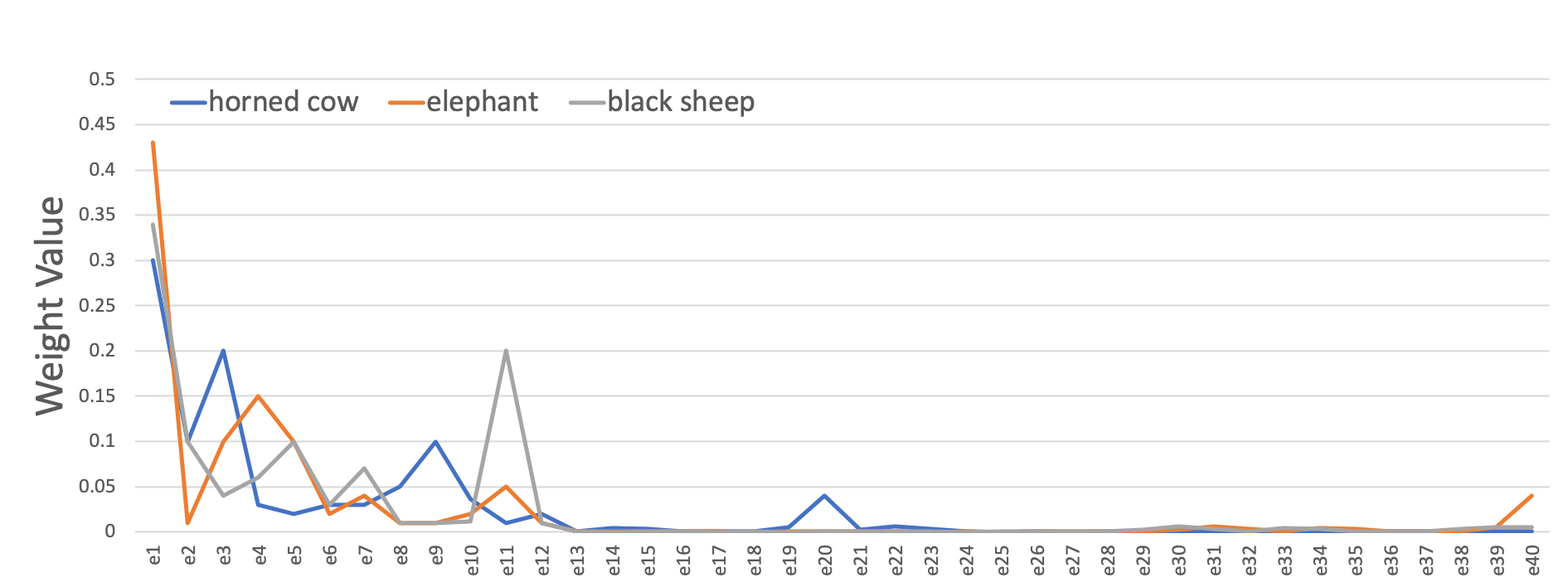}} \\
  \subfloat[]{\includegraphics[width=0.99\linewidth]{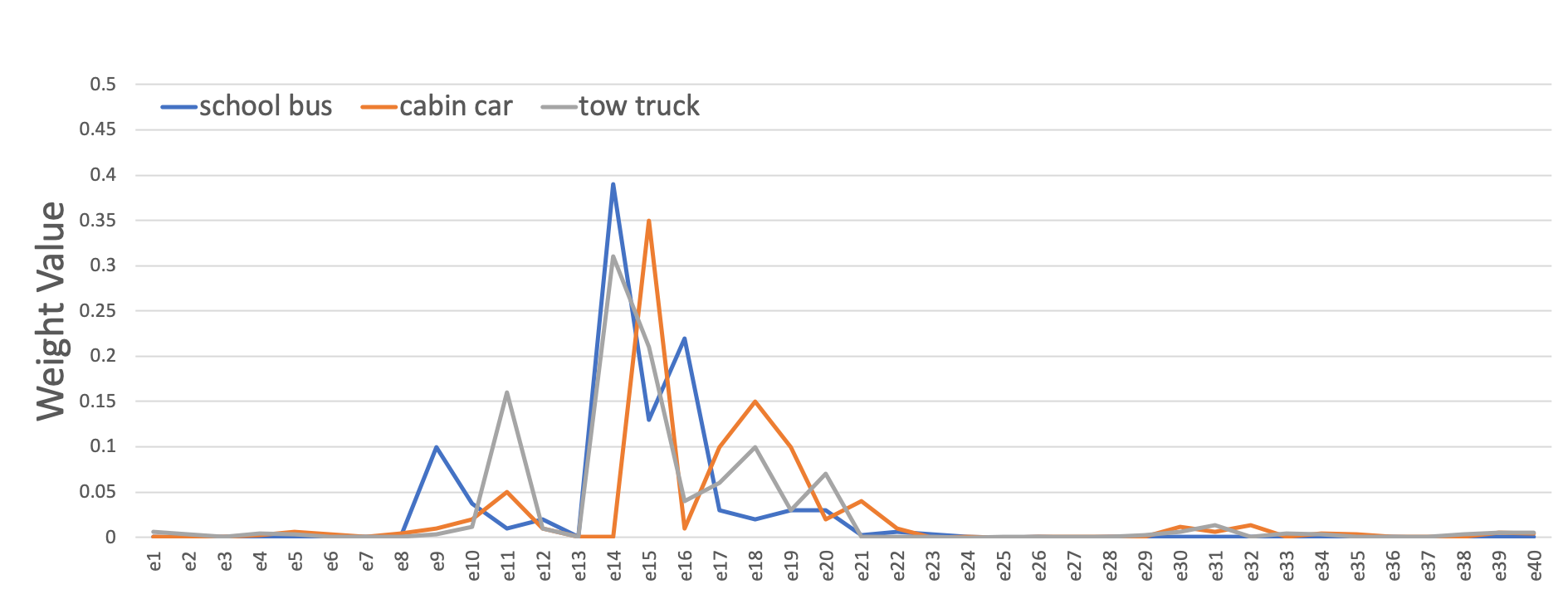}}
  \\
  \subfloat[]{\includegraphics[width=0.99\linewidth]{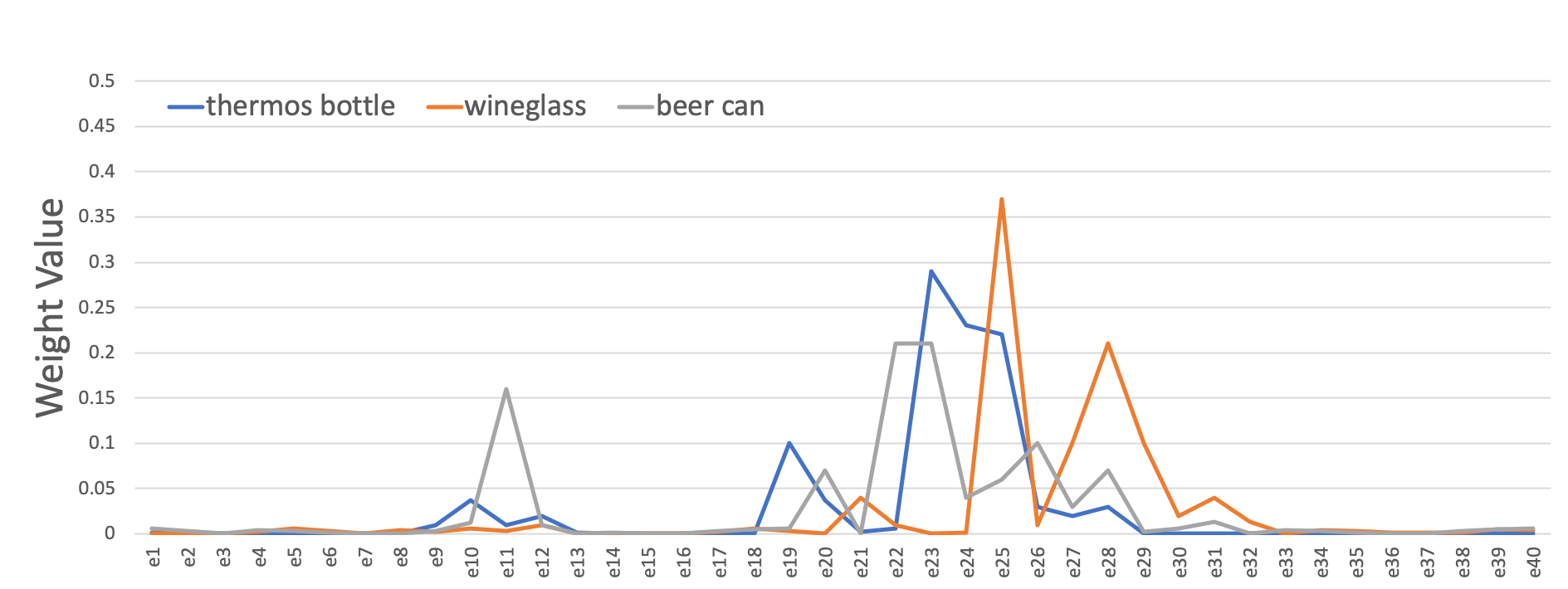}} \\
  \caption{Example aggregation weight distribution. Dynamic aggregation weight on some example object categories of LVIS. The horizontal axis corresponds to the index of expert heads. The vertical axis corresponds to the normalized weight value. The weights are from a CondHead model based on RegionCLIP. The weight indexes are permuted to better show trends of distribution.}
\label{example_weight_distribution}
\end{figure*}

\begin{figure*}[t!]
\centering
\includegraphics[width=0.99\linewidth]{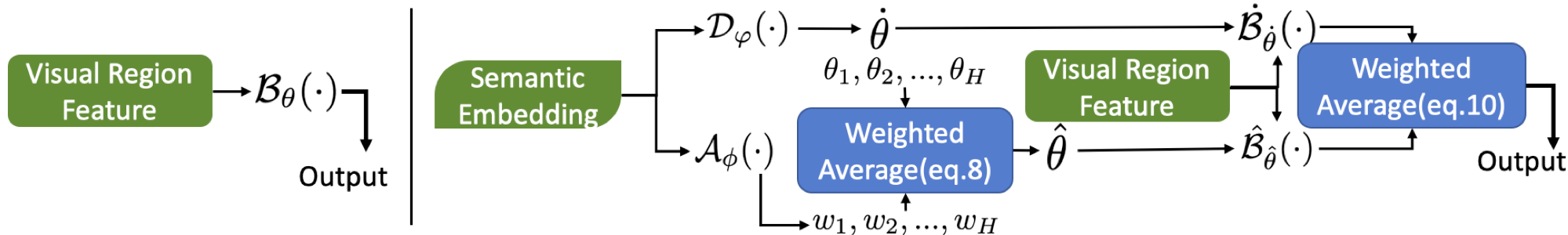}
\caption{\textbf{Left}: illustration of standard box regression head, the learnable parameter is $\theta$. \textbf{Right}: illustration of CondHead architecture, the learnable parameters are $\theta_1, {\theta_2},...,\theta_H, \phi$, and $\varphi$. While box regression ($\mathcal{B}$) is illustrated here, the mask segmentation ($\mathcal{M}$) is similar.}
\label{fig:condhead_arch}
\end{figure*}

\begin{figure*}[t!]
\centering
\includegraphics[width=0.99\linewidth]{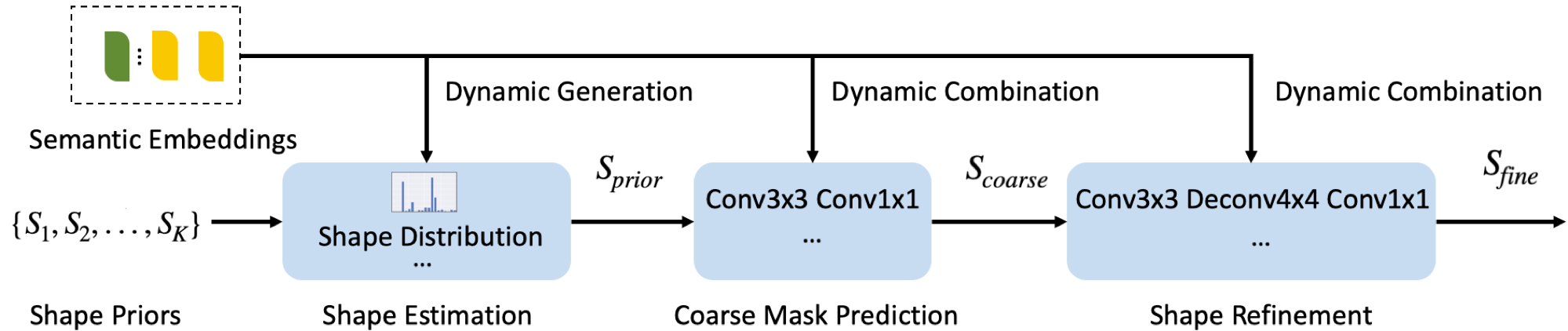}
\caption{Integrating Shapemask into CondHead. We omit the architecture design and only depicts the parametric components that are affected by the proposed CondHead. Dynamic Generation: the shape distribution weights are directly generated on the semantic embedding. Dynamic Combination: the convolution kernels are generated with the proposed dynamically aggregated expert heads.}
\label{fig:shapemask_condhead}
\end{figure*}

\begin{table*}[t!]
\centering
\begin{tabular}{l|ccc|ccc|ccc|ccc}
\toprule
\multirow{3}{*}{} & \multicolumn{6}{c|}{$\mathcal{A}$}                         & \multicolumn{6}{c}{$\mathcal{D}$}                     \\ \cline{2-13}
                  & \multicolumn{3}{c|}{depth} & \multicolumn{3}{c|}{hidden dimension} & \multicolumn{3}{c|}{depth} & \multicolumn{3}{c}{hidden dimension}\\ 
                  & 1     & 2     & 3     & 128     & 256     & 384     & 1     & 2     & 3    & 128    & 256    & 384    \\ \hline
AP$^b_r$                 &   19.0   &  19.9    &   \textbf{20.1}   &   19.3    & \textbf{19.9}     & 19.7     &  19.0     & \textbf{19.9}     & \textbf{19.9}    & 19.5    & \textbf{19.9}    & 19.8   \\
AP$^m_r$                 &  18.7    & \textbf{20.0}     & \textbf{20.0}     & 18.9     & \textbf{20.0}     & 19.7     & 19.4     & 20.0     & \textbf{20.1}    & 18.9    & \textbf{20.0}    & 19.6    \\
\bottomrule
\end{tabular}
\caption{The effect of architectural setting, with dynamic aggregation weight generator ($\mathcal{A}$) and dynamic parameter generator ($\mathcal{D}$). The hidden dimension is fixed at 256 when evaluating the effect of depth. The depth is fixed at 2 when evaluating the effect of hidden dimension. The experiments are conducted on LVIS, with RegionCLIP model based on ResNet-50 backbone.}
\label{arch_ablation_expert_generators}
\end{table*}

\begin{table*}[t!]
\centering
\begin{tabular}{l|ccc|ccc|ccc|ccc}
\toprule
\multirow{3}{*}{} & \multicolumn{6}{c|}{$\mathcal{B}_h$}                         & \multicolumn{6}{c}{$\mathcal{M}_h$}                     \\ \cline{2-13}
                  & \multicolumn{3}{c|}{depth} & \multicolumn{3}{c|}{hidden dimension} & \multicolumn{3}{c|}{depth} & \multicolumn{3}{c}{hidden dimension}\\ 
                  & 1     & 2     & 3     & 128     & 256     & 384     & 2     & 3     & 4    & 128    & 256    & 384    \\ \hline
AP$^b_r$                 &   18.4   &  \textbf{19.6}    &   19.5   &   18.3    & 19.6     & \textbf{19.8}     &  -     & -     & -    & -    & -    & -  \\
AP$^m_r$                 &  -    & -     & -     & -     & -     & -     & 19.4     & \textbf{20.0}     & \textbf{20.0}    & 19.6    & \textbf{20.0}    & 19.9    \\
\bottomrule
\end{tabular}
\caption{The effect of architectural setting, with the expert box regression heads $\{\mathcal{B}_h\}$ and mask segmentation heads $\{\mathcal{M}_h\}$. 
The hidden dimension is fixed at 256 when evaluating the effect of depth. The depth is fixed at 2 when evaluating the effect of hidden dimension. The box heads are instantiated as fully connected networks, and the mask heads are instantiated as convolution networks. The experiments are conducted on LVIS, with RegionCLIP model based on ResNet-50 backbone.}
\vspace{-10pt}
\label{arch_ablation_expert_heads}
\end{table*}

\begin{table}[t!]
\centering
\begin{tabular}{l|ccc|ccc}
\toprule
\multirow{2}{*}{}       & \multicolumn{3}{c|}{$\dot{\mathcal{B}}$} & \multicolumn{3}{c}{$\dot{\mathcal{M}}$} \\ \cline{2-7}
                  & 1     & 2     & 3    & 1     & 2     & 3    \\ \hline
AP$^b_r$                  &   19.9   & \textbf{20.0}     & 19.6    & -     & -     & -    \\
AP$^m_r$                  & -     & -     & -    & \textbf{20.0}     & 19.7     & 19.6    \\ \bottomrule
\end{tabular}
\caption{The effect of architectural setting, with the dynamic generated box regression head $\dot{\mathcal{B}}$, and mask segmentation head $\dot{\mathcal{M}}$. The hidden dimension is fixed at 4 due to the limited output dimension for direct parameter generation.The box heads are instantiated as fully connected networks, and the mask heads are instantiated as convolution networks.
The experiments are conducted on LVIS, with RegionCLIP model based on ResNet-50 backbone.}
\label{arch_ablation_dynamic_heads}
\end{table}

\begin{table}[t]
\centering
\setlength{\tabcolsep}{3pt}
\begin{tabular}{l|cccc|cccc}
\toprule
 & \multicolumn{4}{c|}{Object Detection} & \multicolumn{4}{c}{Instance Segmentation} \\ \hline
 -     & Novel     & Base     & All    & -    & Novel     & Base     & All    & - \\ 
B & 31.3 &    56.5  &   50.4   &  -  &   27.5    &   54.1    &     48.1 &  -   \\
Cd & 33.7 & 58.0 & 52.2 & - & 29.7 & 55.8 & 49.5 & -\\
Cdv & 31.9 & 56.6 & 50.7 & - & 28.4 & 54.7 & 48.5 & -\\ 
Cde & 31.8 & 56.8 & 50.9 & - & 28.2 & 54.6 & 48.4 & -\\\hline
-  & AP$_r$   & AP$_c$   & AP$_f$   & AP   & AP$_r$   & AP$_c$   & AP$_f$   & AP   \\ 
B & 22.1 & 31.8 & 37.0 & 32.4 & 21.8 & 30.2 & 35.1 & 30.2 \\
Cd & 25.1 & 33.4 & 37.8 & 33.9 & 24.4 & 31.6 & 35.9 & 31.6 \\ 
Cdv & 22.5 & 32.1 & 36.9 & 32.5 & 22.1 & 30.6 & 35.3 & 30.5 \\
Cde & 22.9 & 32.5 & 37.2 & 32.8 & 22.5 & 30.6 & 35.4 & 30.8 \\
\bottomrule
 \end{tabular}
\vspace{-4pt}\caption{B, Cd denote baseline and CondHead, Cdv and Cde denote dynamic condition on visual region feature and ensembling/averaging multiple independently trained heads. Results are obtained with RegionCLIP (ResNet50), on COCO and LVIS.}\vspace{-10pt}
\label{table:dynamic_on_visual_feature}
\end{table}

\section*{Example Aggregation Weights}
To analyze how CondHead learns to consolidate the class-wise knowledge into the expert prediction heads, we plot the aggregation weights for the dynamically aggregated head on some example object categories. As shown in Figure~\ref{example_weight_distribution}, we observe evident clustering of the weight distribution on object categories with close semantic meaning. For example, the aggregation weights on \emph{horned cow}, \emph{shepherd dog} and \emph{black sheep} mainly attend to the first 12 expert heads. This is likely because these are all animals and with similar body architecture and pose.
Similarly, we observe the  \emph{school bus}, \emph{cabin car} and \emph{tow truck} attend mainly to the 8th to 22th expert heads (Figure~\ref{example_weight_distribution} (b)). the \emph{thermos bottle}, \emph{wineglass} and \emph{beer can} attend mainly to the 18th to 32th expert heads (Figure~\ref{example_weight_distribution} (c)). 
On the other hand, the detailed weight distribution differ for these highly attended expert heads, this may attribute to the different appearance of these categories. It seems CondHead learns to cope with the difference with compositional knowledge from multiple expert heads.

\section*{More Implementation Details}

\myparagraphonept{Architectural Illustration and Training Details} The proposed CondHead does not involve complex training strategies, it is
simply trained as a straightforward replacement of standard box regression and mask segmentation heads. Fig.~\ref{fig:condhead_arch}
gives an architectural illustration. Concretely, OVR-CNN and RegionCLIP first pre-train the visual-semantic representation and then train open-vocabulary detection by initializing with the learned representation. CondHead is employed in the second-stage training by replacing the original box/mask heads. 
As for ViLD, CondHead is simply trained together with its text-embedding transfer (ViLD-text) and image embedding distillation (ViLD-image), by replacing the original box/mask heads. The text embedding from the CLIP language encoder is used as the semantic embedding for RegionCLIP and ViLD, and the BERT language embedding is used as the semantic embedding for OVR-CNN. Other hyper-parameters such as training schedules follow these baseline methods.

\myparagraphonept{Temperature Annealing} As discussed in Section 3.2 of main paper, we optimize the expert heads with a temperature annealing strategy, \ie, applying large temperature during the early training epochs and gradually annealing the temperature to a small value to ensure good dynamics of Softmax output. Concretely, we set the temperature to 20.0 and linearly decay it to 1.0 during the first 5k iterations.

\myparagraphonept{Integrating Shapemask into CondHead}
Shapemask~\cite{kuo2019shapemask} introduces shape prior and multi-stage refinement to achieve strong class-agnostic instance segmentation, which is validated on the partially supervised instance segmentation task~\cite{hu2018learning}. This prior-based design can be integrated into CondHead to further improve its segmentation quality on open vocabulary objects. 

As shown in Figure~\ref{fig:shapemask_condhead}, we introduce the semantic conditioning on each stage of Shapemask~\cite{kuo2019shapemask}, \ie, shape estimation, coarse mask prediction and shape refinement. Concretely, the shape distribution weights are dynamically generated based on the semantic embedding, the weights are used to average the shape priors to obtain the segmentation prior. Then the two consecutive convolution kernels within the coarse mask prediction module are conditionally parameterized. This is conducted with a set of dynamically combined expert convolution kernels, same as introduced in Section 3.2 of main paper (dynamically aggregated head). Based on the coarse mask prediction, the three consecutive convolution kernels within the shape refinement module are similarly parametrized and utilized to obtain the refined mask segmentation result.

The above are used to instantiate the dynamically aggregated head stream of CondHead (\ie, $\hat{\mathcal{B}}$), the original dynamically generated head stream ($\dot{\mathcal{B}}$) is maintained as introduced in Sec. 3.2.

\section*{The Effect of Architectural Instantiation}
We exam the effect of architectural instantiation on CondHead, specifically with the depth and hidden dimension of the networks. 
\begin{itemize}
    \item Dynamic aggregation weight generator. As shown in Table~\ref{arch_ablation_expert_generators}, the performance improvement is significant when increasing the depth from 1 to 2, while diminishes beyond that, \eg, 0.9 box AP improvement for depth of 1 to 2 and 0.2 box AP improvement for depth of 2 to 3, with dynamic weight generator $\mathcal{A}$. Similar observation holds for the hidden dimension. We thus set the depth and hidden dimension to 2 and 256 for the dynamic weight generator networks.
    \item Expert box regression heads and mask segmentation heads. As shown in Table~\ref{arch_ablation_expert_heads}, for the set of expert heads, the performance improvement is significant when increasing the depth from 1 to 2, while diminishes beyond that, \eg, box AP of 18.4 to 19.6 for depth of 1 to 2 and 19.6 to 19.5 for depth of 2 to 3, Similar observation holds for the hidden dimension. We thus
    set the depth and hidden dimension to 2 and 256 for the expert box heads and mask heads.
    \item Dynamically generated heads. We exam the depth of dynamically generated box and mask heads. Due to the limitation of network output dimension, we set the hidden dimension to 4. As shown in Table~\ref{arch_ablation_dynamic_heads}, increasing the depth actually brings limited benefits, \eg, depth of 1 already achieves 19.9 box AP while depth of 2 and 3 obtain 20.0 and 19.6 box AP. Similar trend is observed with the mask head. We thus employ simple 1-layer network for the dynamic heads.

\end{itemize}

Based on all experiment results above, we set the depth and hidden dimension as shown in the main paper Table 1.

\section*{Ablation on Dynamic Designs}

We also validate the effectiveness of the proposed dynamic conditioning design by examining two other alternative designs: replacing the proposed dynamically combined heads with two simple ensembling baselines (dynamic conditioning on the visual region feature and simply ensembling multiple independently trained heads during inference). As shown in Tab.~\ref{table:dynamic_on_visual_feature}, the performance drops compared to CondHead, especially on novel categories, meaning the proposed method helps better learn class-specific prediction for novel categories.

\end{document}